\documentclass[11pt]{article}

\usepackage{acl2014}
\usepackage{times}
\usepackage{url}
\usepackage{latexsym}
\usepackage{etoolbox}

\usepackage{graphicx}
\usepackage{url}
\usepackage{amsmath}
\usepackage{amssymb}
\usepackage{subfigure}
\usepackage{color}
\usepackage{caption}
\usepackage{multirow}
\usepackage{comment}
\usepackage[utf8]{inputenc}

\usepackage{booktabs}

\newcommand{\bmx}[0]{\begin{bmatrix}}
\newcommand{\emx}[0]{\end{bmatrix}}
\newcommand{\qt}[1]{\left<#1\right>}

\newcommand{\vect}[1]{\mathbf{#1}}

\newcommand{\matr}[1]{\mathbf{#1}}

\newcommand{\vc}[0]{\vect{c}}
\newcommand{\vh}[0]{\vect{h}}

\newcommand{\vx}[0]{\vect{x}}
\newcommand{\vw}[0]{\vect{w}}

\newcommand{\vf}[0]{\vect{f}}
\newcommand{\ve}[0]{\vect{e}}

\newcommand{\mW}[0]{\matr{W}}

\newcommand{\mU}[0]{\matr{U}}

\DeclareMathOperator*{\argmax}{arg\,max}

\graphicspath{ {./figures/} }

\title{Overcoming the Curse of Sentence Length for Neural Machine
Translation using Automatic Segmentation}

\author{
    Jean Pouget-Abadie\thanks{\hspace{2mm}Research done while these authors were visiting Universit\'e de
    Montr\'eal}\\
    Ecole Polytechnique, France \\
  \And
    Dzmitry Bahdanau \footnotemark[1] \\
    Jacobs University, Germany \\
  \AND
    Bart van Merri\"enboer~~~~~~~~~~Kyunghyun Cho \\
    Universit\'e de Montr\'eal, Canada
  \And
    Yoshua Bengio \\
    Universit\'e de Montr\'eal, Canada \\ CIFAR Senior Fellow
}

\date{}

\begin{document}
\maketitle

\begin{abstract}
    The authors of \cite{Cho2014a} have shown that the recently introduced
    neural network translation systems suffer from a significant drop in
    translation quality when translating long sentences, unlike existing
    phrase-based translation systems. In this paper, we propose a way to
    address this issue by automatically segmenting an input sentence into
    phrases that can be easily translated by the neural network translation
    model. Once each segment has been independently translated by the neural
    machine translation model, the translated clauses are concatenated to form
    a final translation. Empirical results show a significant improvement in
    translation quality for long sentences.
\end{abstract}

\section{Introduction}

Up to now, most research efforts in statistical machine translation (SMT)
research have relied on the use of a phrase-based system as suggested
in~\cite{Koehn2003}. Recently, however, an entirely new, neural network based
approach has been proposed by several research groups
\cite{Kalchbrenner2012,Sutskever2014,Cho2014}, showing promising results, both
as a standalone system or as an additional component in the existing
phrase-based system.  In this neural network based approach, an encoder
`encodes' a variable-length input sentence into a fixed-length vector and a
decoder `decodes' a variable-length target sentence from the fixed-length
encoded vector.

It has been observed in \cite{Sutskever2014}, \cite{Kalchbrenner2012} and
\cite{Cho2014a} that this neural network approach works well with short
sentences (e.g., $\lessapprox$ 20 words), but has difficulty with long
sentences (e.g., $\gtrapprox$ 20 words), and particularly with sentences that
are longer than those used for training. Training on long sentences is
difficult because few available training corpora include sufficiently many long
sentences, and because the computational overhead of each update iteration in
training is linearly correlated with the length of training sentences.
Additionally, by the nature of encoding a variable-length sentence into a
fixed-size vector representation, the neural network may fail to encode all the
important details.

In this paper, hence, we propose to translate sentences piece-wise.  We segment
an input sentence into a number of short clauses that can be confidently
translated by the model.  We show empirically that this approach improves
translation quality of long sentences, compared to using a neural network to
translate a whole sentence without segmentation.

\section{RNN Encoder--Decoder for Translation}

The RNN Encoder--Decoder (RNNenc) model is a recent implementation of the
encoder--decoder approach, proposed independently in \cite{Cho2014} and in
\cite{Sutskever2014}. It consists of two RNNs, acting respectively as encoder
and decoder. Each RNN maintains a set of hidden units that makes an `update'
decision for each symbol in an input sequence. This decision depends on the
input symbol and the previous hidden state. The RNNenc in \cite{Cho2014} uses a
special hidden unit that adaptively forgets or remembers the previous hidden
state such that the activation of a hidden unit $h_j^{\qt{t}}$ at time $t$ is
computed by
\begin{align*}
    h_j^{\qt{t}} = z_j h_j^{\qt{t-1}} + (1 - z_j) \tilde{h}_j^{\qt{t}},
\end{align*}
where
\begin{align*}
    \tilde{h}_j^{\qt{t}} =& f\left(
    \left[ \mW \vx \right]_j + r_j \left[ \mU \vh_{\qt{t-1}} \right]
    \right), \\
    z_j =& \sigma\left( \left[ \mW_z \vx \right]_j + \left[\mU_z
    \vh_{\qt{t-1}}\right]_j \right), \\
    r_j =& \sigma\left( \left[ \mW_r \vx \right]_j + \left[\mU_r
    \vh_{\qt{t-1}}\right]_j \right).
\end{align*}
$z_j$ and $r_j$ are respectively the update and reset gates.

Additionally, the RNN in the decoder computes at each step the conditional
probability of a target word:
\begin{multline}
    \label{eq:softmax}
    p(f_{t,j} = 1 \mid f_{t-1}, \dots, f_1, \vc) = \\
    \frac{\exp \left(
        \vw_j \vh_{\qt{t}}\right) } {\sum_{j'=1}^{K} \exp \left( \vw_{j'}
        \vh_{\qt{t}}\right) },
\end{multline}
where $f_{t,j}$ is the indicator variable for the $j$-th word in the target
vocabulary at time $t$ and only a single indicator variable is on ($=1$) each
time. $\vc$ is the context vector, the representation of the input sentence as
encoded by the encoder.  

Although the model in \cite{Cho2014} was originally trained on phrase pairs, it
is straightforward to train the same model with a bilingual, parallel corpus
consisting of sentence pairs as has been done in \cite{Sutskever2014}. In the
remainder of this paper, we use the RNNenc trained on English--French sentence
pairs~\cite{Cho2014a}.

\section{Automatic Segmentation and Translation}
\label{sec:segmentation}

One hypothesis explaining the difficulty encountered by the RNNenc model when
translating long sentences is that a plain, fixed-length vector lacks the
capacity to encode a long sentence. When encoding a long input sentence, the
encoder may lose the track of all the subtleties in the sentence. Consequently,
the decoder has difficult time recovering the correct translation from the
encoded representation. One solution would be to build a larger model with a
larger representation vector to increase the capacity of the model at the price
of higher computational cost.

In this section, however, we propose to segment an input sentence such that each
segmented clause can be easily translated by the RNN Encoder--Decoder. In other
words, we wish to find a segmentation that maximizes the total {\it confidence
score} which is a sum of the confidence scores of the phrases in the
segmentation.  Once the confidence score is defined, the problem of finding the
best segmentation can be formulated as an integer programming problem.

Let $\ve=\left(e_{1},\cdots,e_{n}\right)$ be a source sentence composed of words
$e_{k}$. We denote a phrase, which is a subsequence of $\ve$, with $\ve_{ij} =
\left(e_{i},\cdots,e_{j}\right)$. 

We use the RNN Encoder--Decoder to measure how confidently we can translate a
subsequence $\ve_{ij}$ by considering the log-probability $\log
p(\vf^k\mid\ve_{ij})$ of a candidate translation $\vf^k$ generated by the model.
In addition to the log-probability, we also use the log-probability $\log
p(\ve_{ij} \mid \vf^k)$ from a reverse RNN Encoder--Decoder (translating from a
target language to source language). With these two probabilities, we define the
confidence score of a phrase pair $(\ve_{ij}, \vf^k)$ as:
\begin{multline}
    \label{eq:score}
    c(\ve_{ij},\vf^k) = \frac{\log p(\vf^k \mid \ve_{ij}) + \log q(\ve_{ij}
\mid\vf^k)}{2 \left|\log(j-i + 1)\right|},
\end{multline}
where the denominator penalizes a short segment whose probability is known to
be overestimated by an RNN~\cite{Graves2013}.

The confidence score of a source phrase only is then defined as 
\begin{align}
    \label{eq:confidence}
    c_{ij} = \max\limits_{k} c(\ve_{ij}, \vf_{k}).
\end{align}
We use an approximate beam search to search for the candidate translations
$\vf^k$ of $\ve_{ij}$, that maximize log-likelihood $\log
p(\vf^k|\ve_{ij})$~\cite{Graves13speech,Boulanger2013}.

Let $x_{ij}$ be an indicator variable equal to 1 if we include a phrase
$\ve_{ij}$ in the segmentation, and otherwise, 0. We can rewrite the
segmentation problem as the optimization of the following objective function:
\begin{align}
\label{eq:optimization_objective}
& \underset{\vx}{\text{max}}
& & \sum_{i \leq j} c_{ij} x_{ij} = \vx \cdot \vc \\
& \text{subject to}
& &~\forall k, n_k = 1  \nonumber
\end{align}
$n_k = \sum\limits_{i,j} x_{ij} \mathbf{1}_{i \leq k \leq j}$ is the
number of source phrases chosen in the segmentation containing word $e_k$.

The constraint in Eq.~\eqref{eq:optimization_objective} states that for each
word $\ve_{k}$ in the sentence one and only one of the source phrases contains
this word, $(\ve_{ij})_{i\leq k \leq j}$, is included in the segmentation. The
constraint matrix is totally unimodular making this integer programming problem
solvable in polynomial time.

Let $S^k_j$ be the first index of the $k$-th segment counting from the last
phrase of the optimal segmentation of subsequence $\ve_{1j}$ ($S_j := S^1_j$),
and $s_j$ be the corresponding score of this segmentation ($s_0:=0$). Then, the
following relations hold: 
\begin{align}
 s_{j} =& \max_{1 \leq i \leq j} (c_{ij} + s_{i-1}), &\forall j \geq 1 
 \label{eq:cost}
 \\ 
 S_{j} =&\argmax_{1 \leq i \leq j} (c_{ij} + s_{i-1}), &\forall j \geq 1 
 \label{eq:prev}
\end{align}

With Eq.~\eqref{eq:cost} we can evaluate $s_j$ incrementally. With the evaluated
$s_j$'s, we can compute $S_j$ as well (Eq.~\eqref{eq:prev}).  By the definition
of $S^k_j$ we find the optimal segmentation by decomposing $\ve_{1n}$ into
$\ve_{S^{\overline{k}}_n,S^{\overline{k}-1}_{n} - 1}, \cdots, \ve_{S^2_n,S^1_n -
1}, \ve_{S^1_n, n}$, where $\overline{k}$ is the index of the first one in the
sequence $S^k_n$.  This approach described above requires quadratic time with
respect to sentence length.

\subsection{Issues and Discussion}

The proposed segmentation approach does not avoid the problem of reordering
clauses.  Unless the source and target languages follow roughly the same order,
such as in English to French translations, a simple concatenation of translated
clauses will not necessarily be grammatically correct.

Despite the lack of long-distance reordering\footnote{
    Note that, inside each clause, the words are reordered automatically when
    the clause is translated by the RNN Encoder--Decoder. 
}
in the current approach, we find
nonetheless significant gains in the translation performance of neural machine
translation.  A mechanism to reorder the obtained clause translations is,
however, an important future research question.

Another issue at the heart of any  purely neural machine translation is the
limited model vocabulary size for both source and target languages. As shown in
\cite{Cho2014a}, translation quality drops considerably with just a few unknown
words present in the input sentence.  Interestingly enough, the proposed
segmentation approach appears to be more robust to the presence of unknown words
(see Sec.~\ref{sec:result}). One intuition is that the segmentation leads to
multiple short clauses with less unknown words, which leads to more stable
translation of each clause by the neural translation model.

Finally, the proposed approach is computationally expensive as it requires
scoring all the sub-phrases of an input sentence.  However, the scoring process
can be easily sped up by scoring phrases in parallel, since each phrase can be
scored independently.

\section{Experiment Settings}

\subsection{Dataset}

We evaluate the proposed approach on the task of English-to-French translation.
We use a bilingual, parallel corpus of 348M words selected by the method of
\cite{Axelrod2011} from a combination of Europarl (61M), news commentary (5.5M),
UN (421M) and two crawled corpora of 90M and 780M words respectively.\footnote{
    The datasets and trained Moses models can be downloaded from
    \url{http://www-lium.univ-lemans.fr/~schwenk/cslm_joint_paper/} and the
website of ACL 2014 Ninth Workshop on Statistical Machine Translation (WMT 14).}
The performance of our models was tested on {\tt news-test2012}, {\tt
news-test2013}, and {\tt news-test2014}. When comparing with the phrase-based
SMT system Moses \cite{Koehn2007}, the first two were used as a development set
for tuning Moses while {\tt news-test2014} was used as our test set.

To train the neural network models, we use only the sentence pairs in the
parallel corpus, where both English and French sentences are at most 30 words
long. Furthermore, we limit our vocabulary size to the 30,000 most frequent
words for both English and French. All other words are considered unknown and
mapped to a special token ($\left[ \text{UNK} \right]$).

In both neural network training and automatic segmentation, we do not
incorporate any domain-specific knowledge, except when tokenizing the original
text data.

\subsection{Models and Approaches}
\label{sec:Models and Approaches}

We compare the proposed segmentation-based translation scheme against the same
neural network model translations without segmentation. The neural machine
translation is done by an RNN Encoder--Decoder (RNNenc)~\cite{Cho2014} trained
to maximize the conditional probability of a French translation given an English
sentence. Once the RNNenc is trained, an approximate beam-search is used to find
possible translations with high likelihood.\footnote{
    In all experiments, the beam width is 10.
} 

This RNNenc is used for the proposed segmentation-based approach together with
another RNNenc trained to translate from French to English. The two RNNenc's are
used in the proposed segmentation algorithm to compute the confidence score of
each phrase (See Eqs.~\eqref{eq:score}--\eqref{eq:confidence}). 

We also compare with the translations of a conventional phrase-based machine
translation system, which we expect to be more robust when translating long
sentences.

\section{Results and Analysis}
\label{sec:result}

\subsection{Validity of the Automatic Segmentation}

We validate the proposed segmentation algorithm described in
Sec.~\ref{sec:segmentation} by comparing against two baseline segmentation
approaches. The first one randomly segments an input sentence such that the
distribution of the lengths of random segments has its mean and variance
identical to those of the segments produced by our algorithm. The second
approach follows the proposed algorithm, however, using a uniform random
confidence score.

\begin{table}[ht]
\centering
\begin{tabular}{lll}
\toprule
Model & Test set  \\
\midrule
No segmentation         & 13.15 \\
Random segmentation     & 16.60 \\
Random confidence score & 16.76  \\
Proposed segmentation   & 20.86 \\
\bottomrule
\end{tabular}
\caption{
BLEU score computed on {\tt news-test2014} for two control experiments.
    Random segmentation refers to randomly segmenting a sentence so that    
    the mean and variance of the segment lengths corresponded to the ones 
    our best segmentation method.
    Random confidence score refers to segmenting a sentence with randomly
generated confidence score for each segment.}
\label{tab:control-exp}
\end{table}

From Table~\ref{tab:control-exp} we can clearly see that the proposed
segmentation algorithm results in significantly better performance. One
interesting phenomenon is that any random segmentation was better than the
direct translation without any segmentation. This indirectly agrees well with
the previous finding in \cite{Cho2014a} that the neural machine translation
suffers from long sentences.


\begin{figure*}[ht]
  \centering
  \begin{minipage}{0.31\textwidth}
      \centering
      \includegraphics[width=1.\columnwidth]{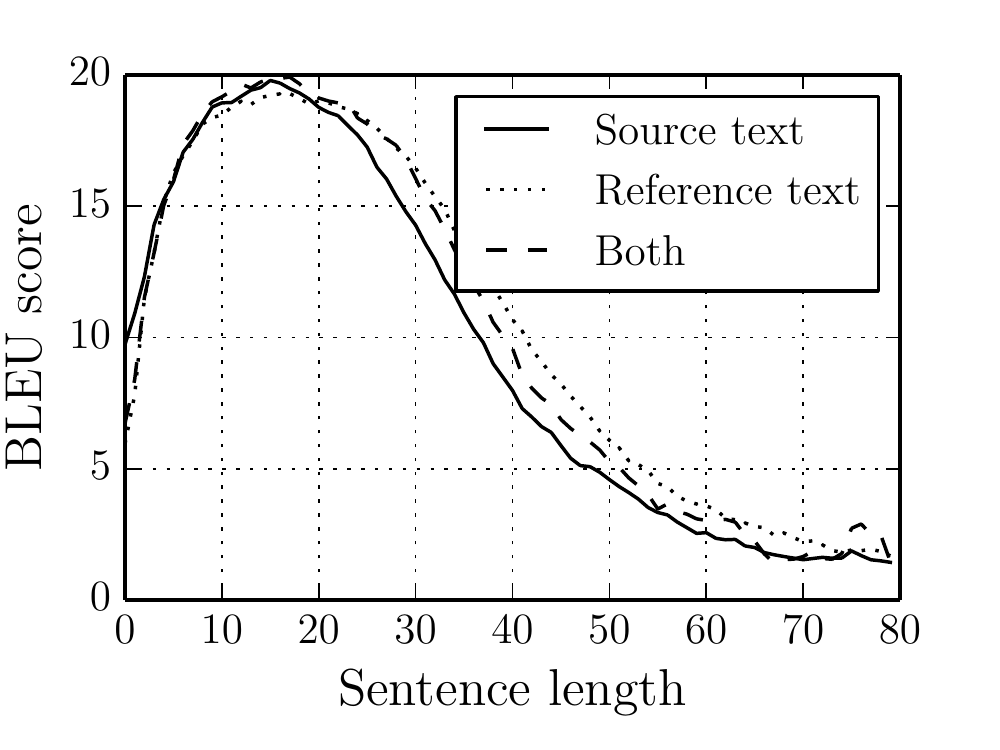}
  \end{minipage}
  \hfill
  \begin{minipage}{0.31\textwidth}
      \centering
      \includegraphics[width=1.\columnwidth]{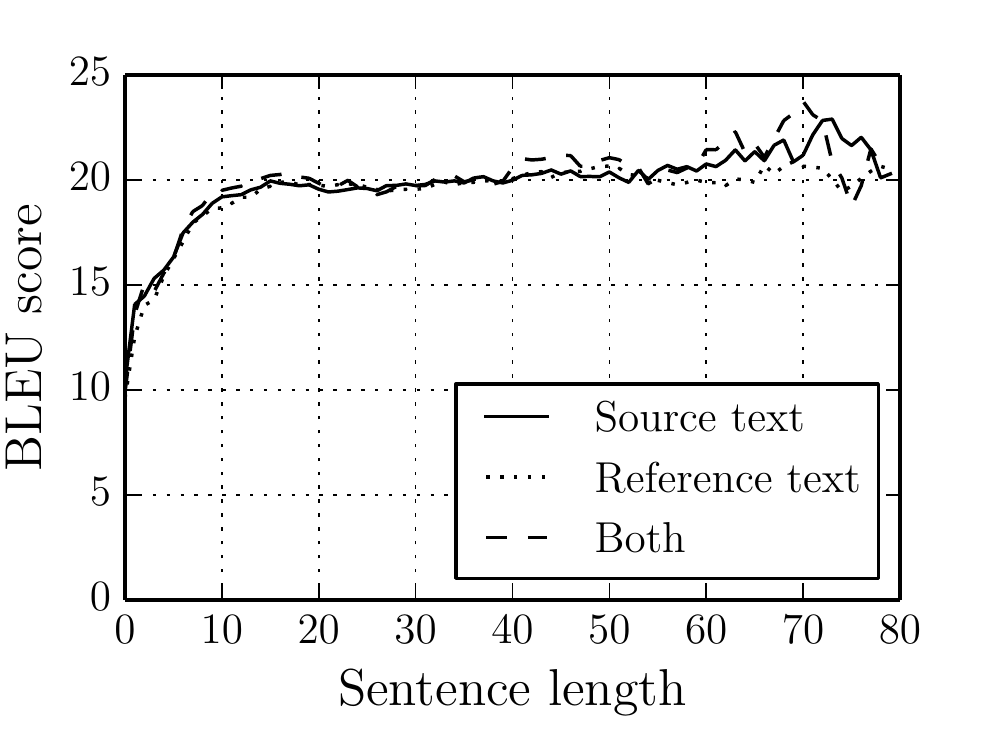}
  \end{minipage}
  \hfill
  \begin{minipage}{0.31\textwidth}
      \centering
      \includegraphics[width=1.\columnwidth]{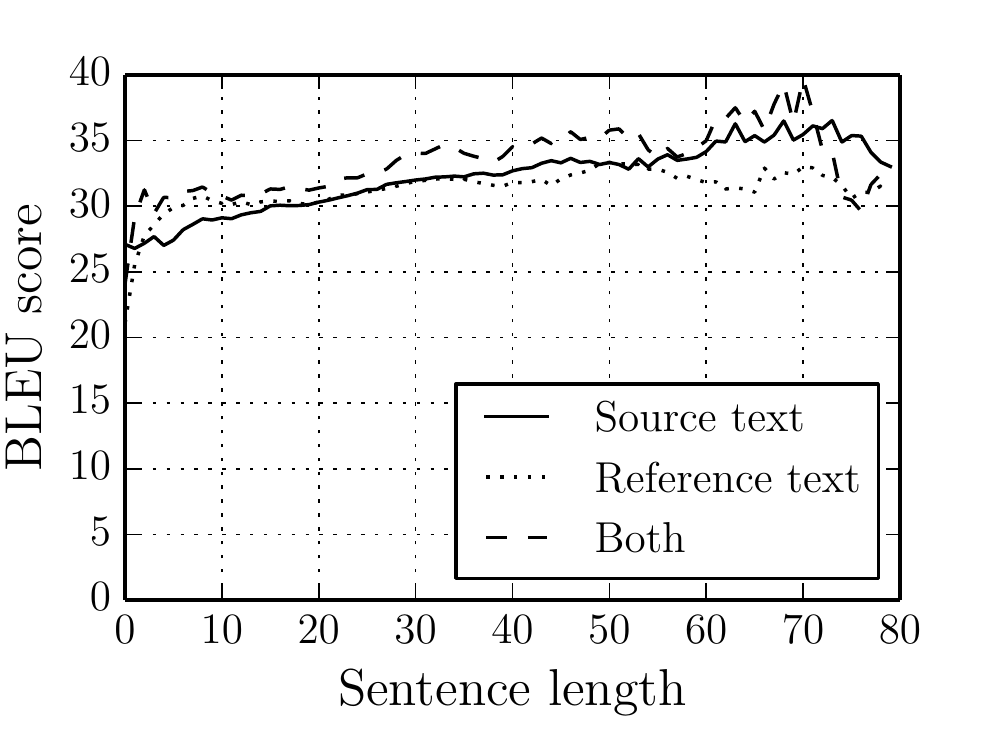}
  \end{minipage}

  \begin{minipage}{0.31\textwidth}
      \centering
      (a) RNNenc without segmentation
  \end{minipage}
  \hfill
  \begin{minipage}{0.31\textwidth}
      \centering
      (b) RNNenc with segmentation
  \end{minipage}
  \hfill
  \begin{minipage}{0.31\textwidth}
      \centering
      (c) Moses
  \end{minipage}
  \caption{The BLEU scores achieved by (a) the RNNenc without segmentation, (b)
  the RNNenc with the penalized reverse confidence score, and (c) the
  phrase-based translation system Moses on a {\tt newstest12-14}.}
  \label{fig:bleu_length}
\end{figure*}

\subsection{Importance of Using an Inverse Model}

\begin{figure}[ht]
  \centering
  \includegraphics[width=1.\columnwidth]{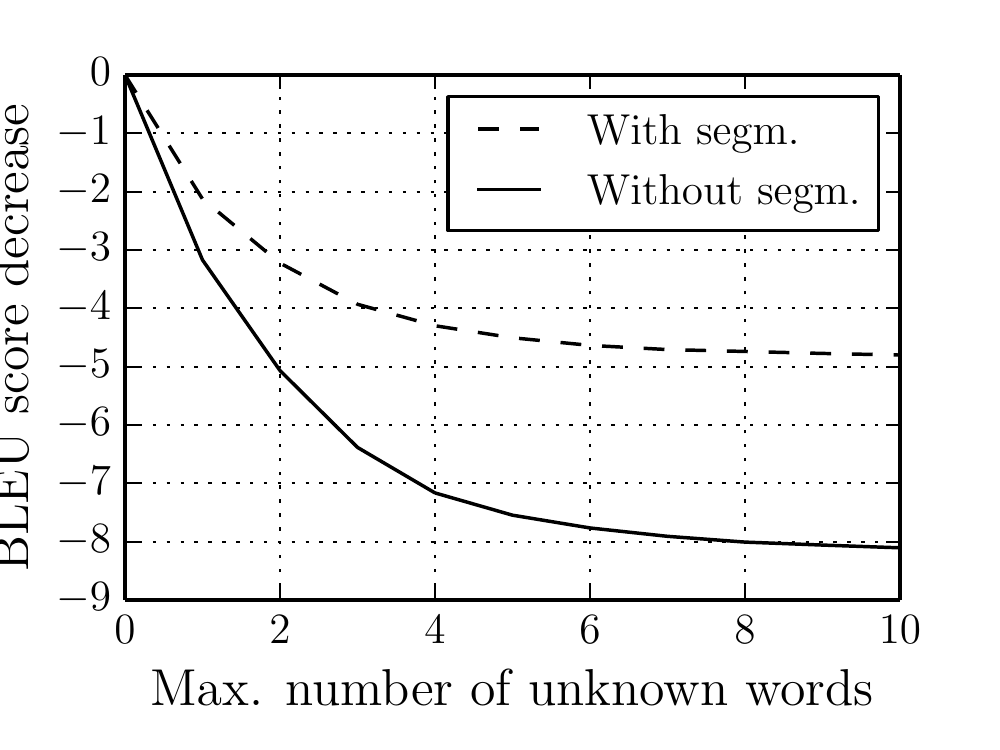}
  \hfill
  \caption{BLEU score loss vs.\ maximum number of unknown words in source and target
sentence when translating with the RNNenc model with and without segmentation.}
  \label{fig:bleu_unk}
\end{figure}

The proposed confidence score averages the scores of a translation model $p(f
\mid e)$ and an inverse translation model $p(e \mid f)$ and penalizes for short
phrases. However, it is possible to use alternate definitions of confidence
score. For instance, one may use only the `direct' translation model or varying
penalties for phrase lengths.

In this section, we test three different confidence score:
\begin{description}
\item[$p(f \mid e)$] Using a single translation model
\item[$p(f \mid e) + p(e \mid f)$] Using both direct and reverse translation
    models without the short phrase penalty
\item[$p(f \mid e) + p(e \mid f)$ (p)] Using both direct and reverse translation
    models together with the short phrase penalty
\end{description}

The results in Table~\ref{tab:bleu} clearly show the importance of using both
translation and inverse translation models. Furthermore, we were able to get the
best performance by incorporating the short phrase penalty (the denominator in
Eq.~\eqref{eq:score}). From here on, thus, we only use the original formulation
of the confidence score which uses the both models and the penalty.

\subsection{Quantitative and Qualitative Analysis}

\begin{table}[ht]
    \centering
    \begin{tabular}{c | c | c c}
        & Model & Dev & Test \\
        \hline
        \hline
        \multirow{5}{*}{\rotatebox[origin=c]{90}{All}}
        & RNNenc                        & 13.15 & 13.92 \\
        & $p(f \mid e)$                 & 12.49 & 13.57 \\
        & $p(f \mid e)+p(e \mid f)$     & 18.82 & 20.10 \\
        & $p(f \mid e)+p(e \mid f)$ (p) & 19.39 & 20.86 \\
        & Moses & 30.64 & 33.30 \\
        \hline
        \multirow{5}{*}{\rotatebox[origin=c]{90}{No UNK}}
        & RNNenc                        & 21.01 & 23.45 \\
        & $p(f \mid e)$                 & 20.94 &  22.62 \\
        & $p(f \mid e)+p(e \mid f)$     & 23.05 & 24.63 \\
        & $p(f \mid e)+p(e \mid f)$ (p) & 23.93 & 26.46 \\
        & Moses  & 32.77 & 35.63 \\
    \end{tabular}
    \caption{BLEU scores computed on the development and test sets. See the text
        for the description of each approach.
        Moses refers to the scores by the conventional phrase-based translation
        system.
    The top five rows consider all sentences of each data set, whilst the
bottom five rows includes only sentences with no unknown words}
    \label{tab:bleu}
\end{table}

As expected, translation with the proposed approach helps significantly with
translating long sentences (see Fig.~\ref{fig:bleu_length}). We observe that
translation performance does not drop for sentences of lengths greater than
those used to train the \mbox{RNNenc} ($\leq$ 30 words).

Similarly, in Fig.~\ref{fig:bleu_unk} we observe that translation quality of the
proposed approach is more robust to the presence of unknown words. We suspect
that the existence of many unknown words make it harder for the RNNenc to
extract the meaning of the sentence clearly, while this is avoided with the
proposed segmentation approach as it effectively allows the RNNenc to deal with
a less number of unknown words.

\begin{table*}[htp]
    \begin{minipage}{0.99\textwidth}
        \small
        \centering
        \begin{tabular}{p{1.6cm} | p{12cm}}

Source&  Between the early 1970s , when the Boeing 747 jumbo defined modern long-haul travel , and the turn of the century , the weight of the average American 40- to 49-year-old male increased by 10 per cent , according to U.S. Health Department Data . \\
\hline
Segmentation&  [[ Between the early 1970s , when the Boeing 747 jumbo defined modern long-haul travel ,] [ and the turn of the century , the weight of the average American 40- to 49-year-old male] [ increased by 10 per cent , according to U.S. Health Department Data .]] \\
\hline
Reference& Entre le début des années 1970 , lorsque le jumbo 747 de Boeing a défini le voyage long-courrier moderne , et le tournant du siècle , le poids de l' Américain moyen de 40 à 49 ans a augmenté de      10 \% , selon les données du département américain de la Santé . \\
\hline
With \mbox{segmentation}& Entre les années 70 , lorsque le Boeing Boeing a défini le transport de voyageurs modernes ; et la fin du siècle , le poids de la moyenne américaine moyenne à l' égard des hommes a augmenté de 10 \% , conformément aux données fournies par le U.S. Department of Health Affairs . \\
\hline
Without segmentation& Entre les années 1970 , lorsque les avions de service Boeing ont dépassé le prix du travail , le taux moyen était de 40 \% . \\
\hline
\multicolumn{2}{c}{} \\
\hline
Source&  During his arrest Ditta picked up his wallet and tried to remove several credit cards but they were all seized and a hair sample was taken fom him. \\
\hline
Segmentation&  [[During his arrest Ditta] [picked up his wallet and tried to remove several credit cards but they were all seized and] [a hair sample was taken from him.]] \\
\hline
Reference& Au cours de son arrestation , Ditta a ramassé son portefeuille et a
tenté de retirer plusieurs cartes de crédit      , mais elles ont toutes été
saisies et on lui a prélevé un échantillon de cheveux . \\
\hline
With \mbox{segmentation}& Pendant son arrestation J' ai utilisé son portefeuille et a essayé de retirer plusieurs cartes de crédit mais toutes les pièces ont été saisies et un échantillon de cheveux a été enlevé. \\
\hline
Without \mbox{segmentation}& Lors de son arrestation  il a tenté de récupérer plusieurs cartes de crédit  mais il a été saisi de tous les coups et des blessures. \\
\hline
\multicolumn{2}{c}{} \\
\hline
Source&  "We can now move forwards and focus on the future and on the 90 \% of assets that make up a really good bank, and on building a great bank for our clients and the United Kingdom," new director general, Ross McEwan, said to the press . \\
\hline
Segmentation&  [["We can now move forwards and focus on the future] [and] [on the 90 \% of assets that make up a really good bank, and on building] [a great bank for our clients and the United Kingdom,"] [new director general, Ross McEwan, said to the press.]] \\
\hline
Reference& "Nous pouvons maintenant aller de l'avant , nous préoccuper de
l'avenir et des 90 \% des actifs      qui constituent une banque vraiment bonne
et construire une grande banque pour la clientèle et pour le Royaume      Uni",
a dit le nouveau directeur général Ross McEwan à la presse . \\
\hline
With \mbox{segmentation}& "Nous pouvons maintenant passer à l'avenir et se concentrer sur l avenir ou sur les 90 \% d actifs qui constituent une bonne banque et sur la construction une grande banque de nos clients et du Royaume-Uni"  Le nouveau directeur général Ross Ross a dit que la presse. \\
\hline
Without \mbox{segmentation}& "Nous pouvons maintenant passer et étudier les 90 \% et mettre en place une banque importante pour la nouvelle banque et le directeur général" a souligné le journaliste . \\
\hline
\multicolumn{2}{c}{} \\
\hline
Source& There are several beautiful flashes - the creation of images has always been one of Chouinard's strong points - like the hair that is ruffled or the black fabric that extends the lines. \\
\hline
Segmentation& [[There are several beautiful flashes - the creation of images has always been one of Chouinard's strong points -] [like the hair that is ruffled or the black fabric that extends the lines.]] \\
\hline
Reference& Il y a quelques beaux flashs - la création d'images a toujours été
une force chez Chouinard - comme ces ch     eveux qui s'ébouriffent ou ces
tissus noirs qui allongent les lignes . \\
\hline
With \mbox{segmentation}& Il existe plusieurs belles images - la création d images a toujours été l un de ses points forts . comme les cheveux comme le vernis ou le tissu noir qui étend les lignes. \\
\hline
Without \mbox{segmentation}& Il existe plusieurs points forts : la création d images est toujours l un des points forts . \\
\hline
\multicolumn{2}{c}{} \\
\hline
Source &  Without specifying the illness she was suffering from, the star performer of `Respect' confirmed to the media on 16 October that the side effects of a treatment she was receiving were `difficult' to deal with. \\
\hline
Segmentation &  [[Without specifying the illness she was suffering from, the star performer of `Respect'] [confirmed to the media on 16 October that the side effects of a treatment she was receiving were] [`difficult' to deal with.]] \\
\hline
Reference& Sans préciser la maladie dont elle souffrait , la célèbre interprète
de Respect avait affirmé aux médias le 16 octobre que les effets
secondaires d'un traitement qu'elle recevait étaient "difficiles". \\
\hline
With \mbox{segmentation} & Sans préciser la maladie qu'elle souffrait  la star de l' `œuvre' de `respect'. Il a été confirmé aux médias le 16 octobre que les effets secondaires d'un traitement ont été reçus. "difficile" de traiter . \\
\hline
Without \mbox{segmentation} & Sans la précision de la maladie  elle a eu l'impression de "marquer le 16 avril' les effets d'un tel `traitement'. \\
        \end{tabular}
    \end{minipage}
    \caption{Sample translations with the RNNenc model taken from the test set
    along with the source sentences and the reference translations.}
    \label{tbl:translations}
\end{table*}

In Table~\ref{tbl:translations}, we show the translations of randomly selected
long sentences (40 or more words). Segmentation improves overall translation
quality, agreeing well with our quantitative result. However, we can also
observe a decrease in translation quality when an input sentence is not
segmented into well-formed sentential clauses.  Additionally, the concatenation
of independently translated segments sometimes negatively impacts fluency,
punctuation, and capitalization by the RNNenc model. Table~\ref{tbl:bad-segm}
shows one such example.

\begin{table*}[htp]
  \begin{minipage}{0.99\textwidth}
    \small
    \centering
    \begin{tabular}{p{2cm} | p{11cm}}
    Source & He nevertheless praised the Government for responding to his request for urgent assistance which he first raised with the Prime Minister at the beginning of May . \\ \hline
    Segmentation & [He nevertheless praised the Government for responding to his request for urgent assistance which he first raised ] [with the Prime Minister at the beginning of May . ]  \\ \hline
    Reference &  Il a néanmoins félicité le gouvernement pour avoir répondu à la demande d' aide urgente qu'il a présentée au Premier ministre début mai . \\ \hline
    With \mbox{segmentation} & Il a néanmoins félicité le Gouvernement de répondre à sa demande d' aide urgente qu'il {\bf a soulevée . avec }le Premier ministre début mai . \\ \hline
    Without \mbox{segmentation}  & Il a néanmoins félicité le gouvernement de répondre à sa demande d' aide urgente qu'il {\bf a adressée au } Premier Ministre début mai . \\
    \end{tabular}
  \end{minipage}
  \caption{An example where an incorrect segmentation negatively
  impacts fluency and punctuation.}
  \label{tbl:bad-segm}
\end{table*}

\section{Discussion and Conclusion}

In this paper we propose an automatic segmentation solution to the `curse of
sentence length' in neural machine translation. By choosing an appropriate
confidence score based on bidirectional translation models, we observed
significant improvement in translation quality for long sentences. 

Our investigation shows that the proposed segmentation-based translation is more
robust to the presence of unknown words.  However, since each segment is
translated in isolation, a segmentation of an input sentence may negatively
impact translation quality, especially the fluency of the translated sentence,
the placement of punctuation marks and the capitalization of words. 

An important research direction in the future is to investigate how to improve
the quality of the translation obtained by concatenating translated segments.

\section*{Acknowledgments}

The authors would like to acknowledge the support of the following agencies for
research funding and computing support: NSERC, Calcul Qu\'{e}bec, Compute Canada,
the Canada Research Chairs and CIFAR.

\bibliographystyle{acl}
\bibliography{strings,strings-shorter,ml,aigaion,myref}

\begin{thebibliography}{}

\bibitem[\protect\citename{Axelrod \bgroup et al.\egroup }2011]{Axelrod2011}
Amittai Axelrod, Xiaodong He, and Jianfeng Gao.
\newblock 2011.
\newblock Domain adaptation via pseudo in-domain data selection.
\newblock In {\em Proceedings of the ACL Conference on Empirical Methods in
  Natural Language Processing (EMNLP)}, pages 355--362. Association for
  Computational Linguistics.

\bibitem[\protect\citename{Boulanger-Lewandowski \bgroup et al.\egroup
  }2013]{Boulanger2013}
Nicolas Boulanger-Lewandowski, Yoshua Bengio, and Pascal Vincent.
\newblock 2013.
\newblock Audio chord recognition with recurrent neural networks.
\newblock In {\em ISMIR}.

\bibitem[\protect\citename{Cho \bgroup et al.\egroup }2014a]{Cho2014a}
Kyunghyun Cho, Bart van Merri\"enboer, Dzmitry Bahdanau, and Yoshua Bengio.
\newblock 2014a.
\newblock On the properties of neural machine translation: {E}ncoder--{D}ecoder
  approaches.
\newblock In {\em Eighth Workshop on Syntax, Semantics and Structure in
  Statistical Translation}, October.

\bibitem[\protect\citename{Cho \bgroup et al.\egroup }2014b]{Cho2014}
Kyunghyun Cho, Bart van Merrienboer, Caglar Gulcehre, Fethi Bougares, Holger
  Schwenk, and Yoshua Bengio.
\newblock 2014b.
\newblock Learning phrase representations using rnn encoder-decoder for
  statistical machine translation.
\newblock In {\em Proceedings of the Empiricial Methods in Natural Language
  Processing (EMNLP 2014)}, October.
\newblock to appear.

\bibitem[\protect\citename{Graves \bgroup et al.\egroup }2013]{Graves13speech}
A.~Graves, A.~Mohamed, and G.~Hinton.
\newblock 2013.
\newblock Speech recognition with deep recurrent neural networks.
\newblock {\em ICASSP}.

\bibitem[\protect\citename{{Graves}}2013]{Graves2013}
A.~{Graves}.
\newblock 2013.
\newblock Generating sequences with recurrent neural networks.
\newblock {\em ar{X}iv:{\tt 1308.0850 [cs.NE]}}, August.

\bibitem[\protect\citename{Kalchbrenner and Blunsom}2013]{Kalchbrenner2012}
Nal Kalchbrenner and Phil Blunsom.
\newblock 2013.
\newblock Two recurrent continuous translation models.
\newblock In {\em Proceedings of the ACL Conference on Empirical Methods in
  Natural Language Processing (EMNLP)}, pages 1700--1709. Association for
  Computational Linguistics.

\bibitem[\protect\citename{Koehn \bgroup et al.\egroup }2003]{Koehn2003}
Philipp Koehn, Franz~Josef Och, and Daniel Marcu.
\newblock 2003.
\newblock Statistical phrase-based translation.
\newblock In {\em Proceedings of the 2003 Conference of the North American
  Chapter of the Association for Computational Linguistics on Human Language
  Technology - Volume 1}, NAACL '03, pages 48--54, Stroudsburg, PA, USA.
  Association for Computational Linguistics.

\bibitem[\protect\citename{Koehn \bgroup et al.\egroup }2007]{Koehn2007}
Philipp Koehn, Hieu Hoang, Alexandra Birch, Chris Callison-Burch, Marcello
  Federico, Nicola Bertoldi, Brooke Cowan, Wade Shen, Christine Moran, Richard
  Zens, Chris Dyer, Ondrej Bojar, Alexandra Constantin, and Evan Herbst.
\newblock 2007.
\newblock Annual meeting of the association for computational linguistics
  (acl).
\newblock Prague, Czech Republic.
\newblock demonstration session.

\bibitem[\protect\citename{Sutskever \bgroup et al.\egroup
  }2014]{Sutskever2014}
Ilya Sutskever, Oriol Vinyals, and Quoc Le.
\newblock 2014.
\newblock Sequence to sequence learning with neural networks.
\newblock In {\em Advances in Neural Information Processing Systems (NIPS
  2014)}, December.

\end{thebibliography}

\end{document}